\documentclass[twoside,leqno,twocolumn]{article}

\usepackage[letterpaper]{geometry}

\usepackage{ltexpprt}
\usepackage{hyperref}

\usepackage{yhmath}
\usepackage{hyperref}
\usepackage{url}
\usepackage{graphicx}
\usepackage{algorithm}
\usepackage{multirow}
\usepackage[switch]{lineno}
\usepackage[noend]{algpseudocode}

\usepackage{makecell}

\begin{document}

\title{\Large Spatiotemporal Classification with limited labels using Constrained Clustering for large datasets}

\author{
    Praveen Ravirathinam\thanks{These authors contributed equally to this article} $^\ddag$\and
    Rahul Ghosh$^{*\ddag}$\and
    Ke Wang$^\ddag$\and
    Keyang Xuan$^\ddag$\and
    Ankush Khandelwal$^\ddag$\and
    Hilary Dugan$^\dagger$\and
    Paul Hanson\thanks{University of Wisconsin-Madison. \{hdugan, pchanson\}@wisc.edu}\and
    Vipin Kumar\thanks{University of Minnesota. \{pravirat, ghosh128, wan00802, xuan0008, khand035, kumar001\}@umn.edu}
}

\date{}

\maketitle



\begin{abstract}
\small\baselineskip=9pt Creating separable representations via representation learning and clustering is critical in analyzing large unstructured datasets with only a few labels. Separable representations can lead to supervised models with better classification capabilities and additionally aid in generating new labeled samples. Most unsupervised and semisupervised methods to analyze large datasets do not leverage the existing small amounts of labels to get better representations. In this paper, we propose a spatiotemporal clustering paradigm that uses spatial and temporal features combined with a constrained loss to produce separable representations. We show the working of this method on the newly published dataset ReaLSAT, a dataset of surface water dynamics for over 680,000 lakes across the world, making it an essential dataset in terms of ecology and sustainability. Using this large unlabelled dataset, we first show how a spatiotemporal representation is better compared to just spatial or temporal representation. We then show how we can learn even better representation using a constrained loss with few labels. We conclude by showing how our method, using few labels, can pick out new labeled samples from the unlabeled data, which can be used to augment supervised methods leading to better classification.
\end{abstract}


\section{Introduction}

With advances in sensing technologies, computing power, and machine learning algorithms, many large-scale data sets are becoming available in scientific domains. However, most of these large-scale datasets either lack or have limited labels, making it hard to build supervised models around them for tasks such as classification or segmentation. One example is a newly published dataset ReaLSAT~\cite{realsat}, a collection of surface water dynamics for over 681,000 water bodies worldwide from 1984 to 2020. Due to its large size and rich information about a wide variety of water bodies, this dataset has high ecological and social importance. ReaLSAT, however, has no labels associated with each water body as to what type of water body it is, making it a sizeable unlabelled dataset. The ability to conclude the function of inland waters globally depends on the ability to identify and classify millions of water bodies. Reservoirs function differently than natural lakes, which differ from rivers and agricultural ponds. Historically, the best classifications were painstakingly mapped in detail by regional authorities, whereas newer datasets are manually quality-controlled through satellite imagery. The state-of-the-art dataset of global inland waters includes minimal classifications and no temporal dynamics~\cite{hydrolakes}, partly because global surface water bodies constitute a large, heterogeneous dataset that is nearly impossible to classify manually. Methodologies developed to classify waterbody types (i.e., reservoirs, ponds, ephemeral lakes, and others) would significantly impact domain research focused on global water resources and climate impacts. In particular, the ability to label ReaLSAT would be beneficial to various communities worldwide as it would enable the community to draw conclusions on the function of inland waters globally.

Deep learning has dramatically advanced the state-of-the-art in classification using remotely sensed data~\cite{calcrop21,penatti2015deep,maggiori2016convolutional}. However, accurately identifying water bodies using remotely sensed data and supervised machine learning faces several key challenges. First, differentiating between types of water bodies using a single modality data is challenging as two classes of water bodies can have common aspects when observed through a particular view of the data. For example, stable lakes and seasonal lakes can be harder to distinguish using remotely sensed images due to similar shapes, whereas rivers and farms show very similar water change patterns and thus are harder to distinguish using these temporal changes. Second, most of these large-scale datasets either have no or limited labels because creating labels requires manual inspection, making it hard to train state-of-the-art deep learning models. This paper presents a framework to address these challenges by leveraging ideas from three different research themes discussed below.

The first research theme involves combining different modalities to improve underlying data representations to boost the performance of supervised/unsupervised models\cite{shi2019variational,alayrac2020self,visualmultimodal}. The intuition for these multimodal approaches is similar to how humans perceive their surroundings by combining multiple modalities such as sight, sound, and smell to understand their environment better\cite{brainmodal}. Another subcategory of work in this theme are approaches that build representations for entire space-time data (e.g., video) \cite{qian2021spatiotemporal}.

The second theme involves using unsupervised, self-supervised, and semi-supervised algorithms~\cite{ji2019invariant,enguehard2019semi,ranzato2007unsupervised,xie2016unsupervised} to build good representations that can lead to improved classification performance. Of these, autoencoder-based methods~\cite{vincent2010stacked} are used most widely as they do not require any labeled data. Deep embedded clustering (DEC)~\cite{xie2016unsupervised} is a recent variation on this theme, which learns representation while optimizing a clustering objective. The intuition behind such approach is that the samples (even though unlabeled) are clustered in some space, which such approaches try to identify. More recent works extend this approach by leveraging labeled data (which is already available as training data in classification setting) to create additional constraint-based loss components (e.g., must-link, cannot-link) \cite{zhang2019framework}. It has been seen across domains that using labelled data along with unlabelled data has led to better results and feature representation \cite{1007592,shi2021semi,goldman2000enhancing}

The third theme involves the clustering-based extraction of new labels from large pools of unlabelled data. A standard method is first to cluster the data and then label entire clusters with a class label if most (or all) of the labels present in the cluster belong to the same class~\cite{selflabellingiclr}.It has also been shown that it is possible to label large datasets in absence of labels using invariant information clustering techniques\cite{ji2019invariant} and methods also show assigning each data point a unique label \cite{alexey2015discriminative,wu2018unsupervised}.

Our proposed framework combines ideas from all three themes to build high-quality classification models even in a small number of labeled samples. The framework is illustrated in the context of our motivating problem (labeling lake types) but is generally applicable. Specifically, we use an autoencoder-based method to build good representations of ReaLSAT by combining the spatial (the monthly water land maps) and temporal (the water count over time) modalities of ReaLSAT to learn robust multi-view representations. Further, we explore how having a small number of labels can boost learned data representation using a constrained loss framework. Additionally, we present how to efficiently extract labels from the pool of unlabelled data using clusters formed by the data representations. We then show how training supervised models using these extracted labels significantly boosts the classification performance over training with few labels.

To summarise our major contributions are 
\begin{enumerate}
    \item We propose a semi supervised spatiotemporal clustering paradigm that uses a small number of labels and a constrained loss to significantly boost representation quality for large scale datasets
    \item We show how to effectively implement constrained loss in the presence of a small number of labels.
    \item We show how one can use our paradigm to extract new labelled samples from large pools of unlabelled data given a small number of labeled samples.
    \item We show how supervised models trained with our paradigm boost overall classification performance on ReaLSAT significantly when compared to competitive methods
    \item We release code and data to be used by the wider scientific community \footnote{\href{https://drive.google.com/drive/folders/11hE37lwpCtUtBaEPPoeU4ELL8BiaxDkL?usp=sharing}{\underline{Code Link}}}
\end{enumerate}

\begin{figure*}[t]
    \centering
    \includegraphics[width = 0.8\textwidth]{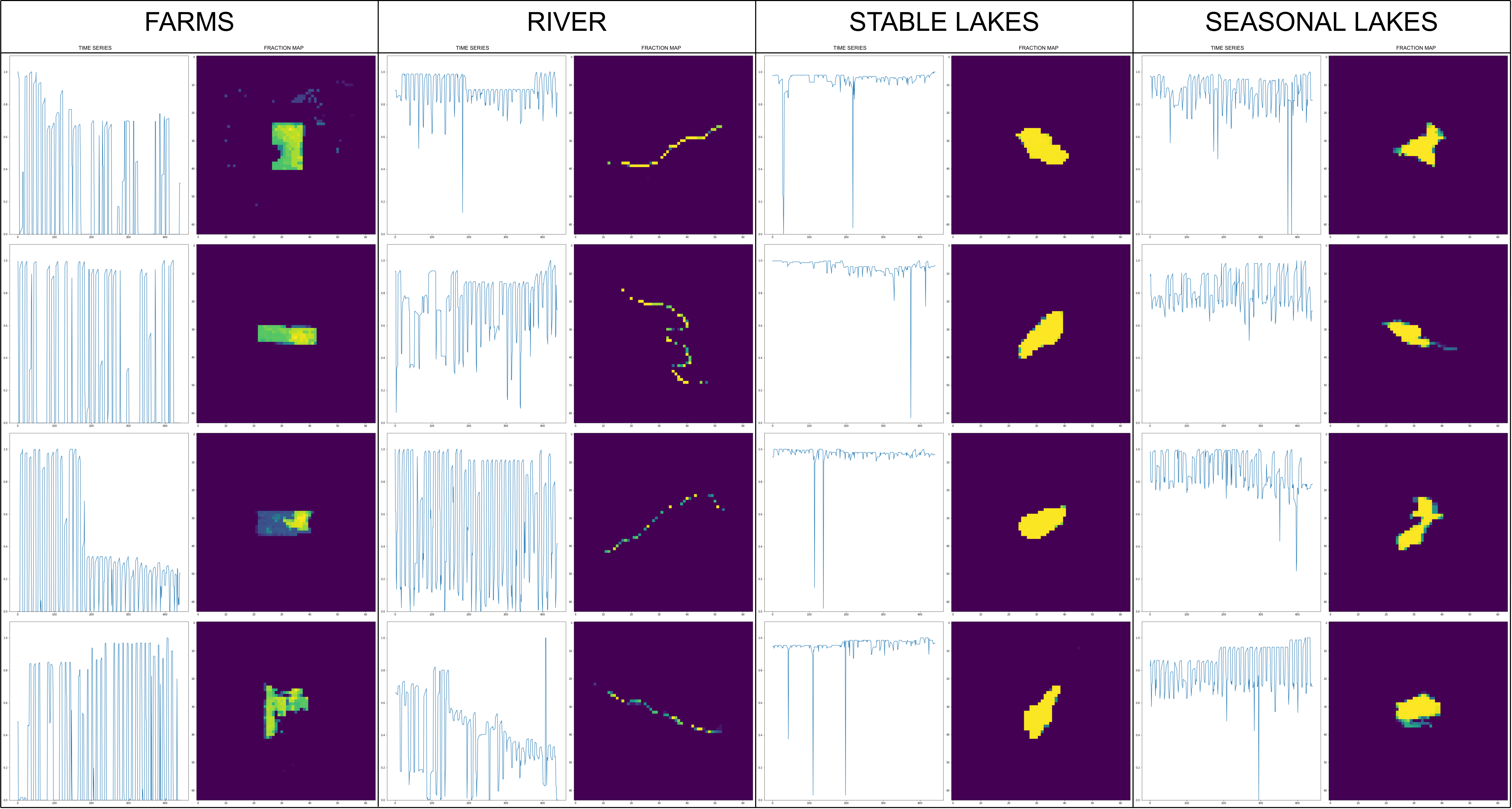}
    \caption{A diagrammatic representation for some water bodies from each class in the two modalities: Fraction Map and Time Series}
    \label{fig:modalities}
\end{figure*}

\begin{figure*}[t]
    \centering
    \includegraphics[width = 0.9 \textwidth]{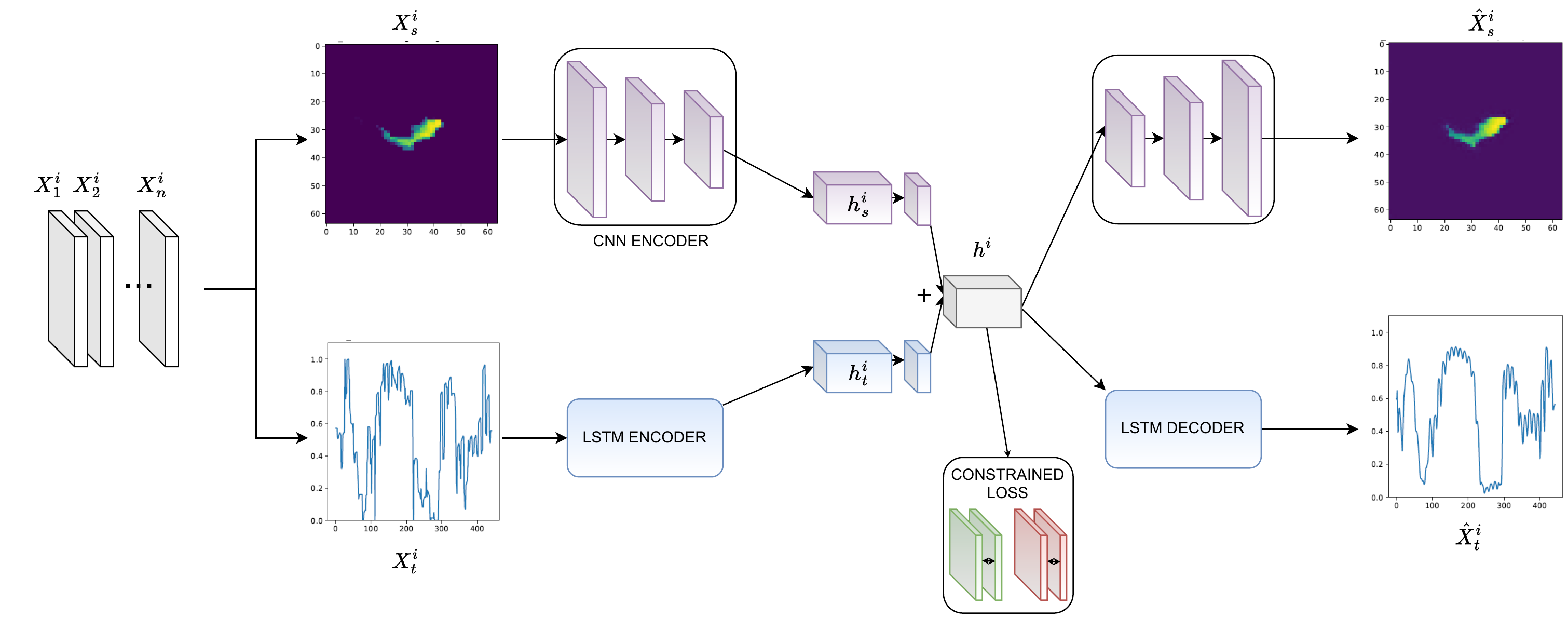}
    \caption{A diagrammatic representation of our proposed semi-supervised autoencoder-based pipeline.}
    \label{fig:architecture}
\end{figure*}

\vspace{-0.2in}
\section{Dataset and Problem Setting}

For this paper, we use the 8642 water bodies in North America that lie within the area range of 0.31 and 0.36 sq km. We analyze four classes, namely, Farm based water bodies (Farms), River segments (River), Natural stable lakes (Stable Lakes), and seasonal lakes (Seasonal Lakes). We manually label close to 1100 water bodies out of the 8642 in our subset. Among these, 417 are labeled as Farms, 316 as River, 142 as Stable Lakes, and 288 are labeled as Seasonal Lakes). We then split the data into two halves such that each class had equal numbers on either half. The unlabeled samples were split at random into equal numbers.

As mentioned, water bodies from the same classes have some common aspects. However, identifying these aspects from the raw spatiotemporal stack available from ReaLSAT can be challenging, which is why we propose a modality-based approach. Specifically, each water body $i$ is denoted by a spatiotemporal stack $X_i$, consisting of a binary map for each month of the year showing whether each pixel is land or water. From this spatiotemporal stack, we extract information corresponding to the two modalities, spatial and temporal. The spatial modality is also called the fraction map $X^i_s$, which is the map generated by taking the mean of all the binary maps present in $X_i$. Each pixel in the resulting fraction map would represent the percentage of timestamps in which that pixel was water across the months. If a pixel has a value of 0, that means that it was never water, implying that this pixel was always labeled as land. Analysis of the non-zero values in the fraction map can give us a sense of shape and which parts of the water body vary, but we cannot extract stability or rapid shifts from this modality. The temporal modality is called the surface area time-series $X^i_t$ of the water body. This modality is constructed by counting the number of water pixels in each binary map and then normalizing with respect to the maximum value. Finally, we are left with a time series of lengths equal to the number of months in ReaLSAT, with each value lying between 0 and 1. Here, if a particular month has a value of 0.9, it means that the water body is 90\% full, and a value of 0 means the water body is empty at that particular month. Using this modality, we can observe rapid changes or periods of stability, but not the shape of the water body.

Before taking the mean image to get a fraction map, we first pad the edges of the image with zeroes to make a square image, then resize it to $64\times64$ using nearest neighbor interpolation. We calculate the time series water count from the original non-resized monthly binary map. In the end, the fraction map shapes are (64,64,1), and the time series shapes are (442,1). Some examples of water bodies from each class in the two respective modalities can be seen in Figure \ref{fig:modalities}.

\floatname{algorithm}{Algorithm}
\renewcommand{\algorithmicrequire}{\textbf{Input:}}
\renewcommand{\algorithmicensure}{\textbf{Output:}}


\section{Method}
This Section provides details of the proposed representation learning with limited labels and the label augmentation approach. The methods presented in this paper aim to tackle three sub-tasks: (i) how to combine information from different modalities of data, (ii) how to learn robust representations by leveraging the constraints from the limited number of labels, and (iii) how to further augment the limited number labels with new labels from unlabeled data for improving supervised classification of inland water bodies. In Section~\ref{Sec:Multimodal Auto-encoder}, we first introduce the multimodal auto-encoder that uses the data from two modalities to represent water bodies. Then in Section~\ref{Sec:Constrained Loss}, we discuss our proposed constrained loss that leverages the limited labels to enforce the similarity of representations of water bodies from the same class. Finally, in Section~\ref{Sec:Label Augmentation}, we describe how to leverage the clustering structure to extract new labels from the unlabeled data and train robust supervised classifiers.

\subsection{Multimodal Auto-encoder}
\label{Sec:Multimodal Auto-encoder}
As discussed, water bodies can be accurately distinguished only when we combine information from different modalities of data. Thus our objective is to build a robust representation of water bodies from the multimodal image and time-series data. This paper uses a multimodal autoencoder-based approach that combines the information by sharing the latent space between the two modalities. This model's encoder ($\mathcal{E}$) and decoder ($\mathcal{D}$) have two components with separate parameters, each for the spatial and temporal data. The fraction map $X^i_s$ of a water body $i$ is fed into the spatial part of the encoder $\mathcal{E}(\cdot, \theta^\mathcal{E}_s)$ to generate the spatial embeddings $\boldsymbol{h}^i_s$. The spatial encoder is a convolutional neural network (CNN) based architecture with $\theta_s$ as the parameters, and the spatial embeddings are essentially the output from the network's last layer. Similarly, the surface area time-series $X^i_t$ of the water body is fed into the temporal part of the encoder $\mathcal{E}(\cdot, \theta\mathcal{E}_t)$ to generate the temporal embeddings $\boldsymbol{h}^i_s$. Our temporal encoder is a bi-directional LSTM as they are particularly suited for our task, where embedding long-term temporal dependencies and the overall water behaviors in a sequence are needed. The embeddings from the forward and backward LSTM are added to get the final temporal embeddings.

The spatial and temporal embeddings have the same dimension and are combined using a combination operator. The combination operator consists of a transformation function parametrized by $\theta^h_s$ and $\theta^h_t$ followed by an activation function and normalization layer. The final multimodal embeddings are obtained from this combination operator, as shown,
\vspace{-0.1cm}
\begin{equation}
    \boldsymbol{h}^i = \frac{\boldsymbol{h}^i_s}{\|\boldsymbol{h}^i_s\|_p} + \frac{\boldsymbol{h}^i_t}{\|\boldsymbol{h}^i_t\|_p}\qquad\text{where,}
    \begin{cases}
        \boldsymbol{h}^i_s &= \alpha(\mathcal{F}(\boldsymbol{h}^i_s, \theta^h_s))) \\
        \boldsymbol{h}^i_t &= \alpha(\mathcal{F}(\boldsymbol{h}^i_t, \theta^h_t)))
    \end{cases}
\end{equation}

We use the ReLU activation function and the $\ell_2$ normalization in our work. In our experiments, our combination operator learns better representations that directly normalize the single modality embeddings and adding. Moreover, not normalizing the single modality embeddings before combining leads to scaling issues and one embedding dominating the other. Finally, the multimodal embedding is fed to both the spatial $\mathcal{D}(\cdot, \theta\mathcal{D}_s)$ and temporal $\mathcal{D}(\cdot, \theta\mathcal{D}_t)$ part of the decoder. This design choice forces the autoencoder to share the same latent space for both modalities and, thus, an effective combination of information. The spatial decoder is CNN based decoder consisting of up-convolutional layers and creates the reconstructed fraction map $\hat{X}^i_s$. At the same time, the temporal decoder consists of an LSTM-based generator to obtain the reconstructed surface area time series $\hat{X}^i_t$. The temporal decoder iteratively outputs the data at each time based on the output data from the previous time steps.

The parameters of this multimodal autoencoder are trained to minimize the reconstruction loss computed as the mean-squared error between the reconstructed and the original inputs. The two-part reconstruction loss is given by the Fraction Map reconstruction error ${L}^{Rec}_s$ and the surface area reconstruction error ${L}^{Rec}_t$, as shown,
\vspace{-0.5cm}
\begin{equation}
\label{eq:reconstruction_loss}
    \begin{split}
        \mathcal{L}^{Rec}_s &= \frac{1}{B}\sum_{\forall{h,w}} (X^i_s [h,w] - \hat{X}^i_s [h,w])^2\\
        \mathcal{L}^{Rec}_t &= \frac{1}{B}\sum_{\forall{n}} (X^i_t [n] - \hat{X}^i_t [n])^2\\
    \end{split}
\end{equation}
where, B is the number of images, n in number of timestamps, (h,w) are the height and the widths of the images respectively 

\subsection{Constrained Loss}
\label{Sec:Constrained Loss}
Our proposed constrained loss is based on the fundamental idea that samples of same class should have similar representations in the latent space. In this setting, if we know two water bodies $i$ and $j$ are of the same class, then their multimodal representations $\boldsymbol{h}^i$ and $\boldsymbol{h}^j$ should be similar. However, choosing a particular similarity function is a critical design choice. We use the cosine between the two embeddings as the similarity measure. This ensures the embeddings of the two water bodies belonging to the same class lie in the same direction, i.e., the angle between the vectors should tend to 0. Our constrained loss is thus defined as,
\vspace{-0.7cm}

\begin{equation}
\label{cl_eq}
    \mathcal{L}^{Constrained} = \frac{1}{\text{N}}\sum_{\forall{classes}}\sum_{\forall{pairs}} - \log(\frac{|\boldsymbol{h}^i\cdot\boldsymbol{h}^j|}{|\boldsymbol{h}^i||\boldsymbol{h}^j|})
\end{equation}

where, N is number\_classes * number\_pairs\_per\_class. Our total constrained loss is the sum of all such logarithms we obtain from each water body pair given for each class. In our experiment, we find that using cosine as the similarity leads to smoother convergence and better representations. Using euclidean distance can lead to exploding loss values as well as it enforces stronger constraints which have the potential to lead to trivial solutions. Additionally, the reconstruction loss prevents the model from collapsing to a degenerate solution by ensuring that the decoder can reconstruct the data point using the embeddings. 

Combining the constrained loss~\ref{cl_eq} and the reconstruction loss~\ref{eq:reconstruction_loss}, we get the final training loss as follows:
\begin{equation}
\label{total_loss}
    \mathcal{L} = \lambda \mathcal{L}^{Rec}_s +  \gamma \mathcal{L}^{Rec}_t + \mathcal{L}^{Constrained}
\end{equation}

where, $\lambda$ and $\gamma$ are the hyper-parameters balancing the contributions of each of the losses. Due to our spatial and temporal components along with a unique constrained loss (given by Eq. \ref{cl_eq}), we call our paradigm SpatiaL plus TemporaL AutoEncoder with Constained Loss, SLTLAE\_CL. Note that this approach from another recent approach that uses constrained loss \cite{zhang2019framework} in three significant ways: i) SLTLAE\_CL does not use cannot link constrains and ii) SLTLAE\_CL uses a joint loss as opposed to alternating the reconstruction and constrained losses while training. iii) SLTLAE\_CL does not have a deep embedded clustering component. In the evaluation section, we compare the performance of our proposed approach with this scheme (referred to as SLTLAE\_CONS) and another scheme based in deep embedded clustering \cite{xie2016unsupervised} (referred to as SLTLAE\_DEC).

\begin{algorithm}[t]
    \caption{Training Algorithm}\label{alg:cap}
    \begin{algorithmic}
        \Require Fraction map of size 64 x 64 x 1, Time Series of shape 442 x 1, Epochs M,  Batches B, Classes C
        \Ensure Fraction map of size 64 x 64 x 1, Time Series of shape 442 x 1, latent representations $ID_{enc}s$
        \\\\
        Initialize the pipeline with weights  
        \For {epoch = 1 to $M$}
            \If{epoch $\leq$ M/2}
                \For {batch = 1 to B} 
                    \State {\scriptsize Calculate Total Loss $L = \lambda \mathcal{L}^{Rec}_s +  \gamma \mathcal{L}^{Rec}_t$}
                    \State {\scriptsize Update weights based on $L$}
                \EndFor
            \EndIf
            \If{epoch $>$ M/2}
                \For {batch = 1 to B} 
                    \State {\scriptsize $\mathcal{L}^{Constrained} = 0$}
                    \State {\scriptsize Get per class count in batch $B_i \rightarrow (C_1,C_2..C_C)$}
                    \State {\scriptsize Get minimum Class count $C_{min}$ from  $(C_1,C_2..C_C)$}
                    \For{Class = $C_1$ to $C_C$}
                        \State {\scriptsize Create $2*C_{min}$ pairs from samples labeled as $C_i$}
                        \State {\scriptsize Store these $2*C_{min}$ pairs in $Pairs\_Class_{Ci}$}
                        \State {\scriptsize Calculate $\mathcal{L}^{Constrained\_Ci}$ using $Pairs\_Class_{Ci}$ }
                        \State {\scriptsize Append $\mathcal{L}^{Constrained\_Ci}$ to $\mathcal{L}^{Constrained}$}
                    \EndFor
                    \State {\scriptsize $\mathcal{L}^{Constrained}$ = $\mathcal{L}^{Constrained}$/$C$}
                    \State {\scriptsize Calculate Total Loss $L$ from Eq. \ref{total_loss}}
                    \State {\scriptsize Update weights based on $L$}
                \EndFor
            \EndIf
        \EndFor
    \end{algorithmic}

\end{algorithm}

{\bf{Implementation details}}:
We found a batch size of 256, $\lambda = 0.01$ and $\gamma = 1$ were the best fit for our purpose. We also found that it is critical to make sure that one passes an equal number of pairs from each class for $L_{Constrained}$. Passing more pairs of a certain class led to bias in minimizing the constrained loss of that class, implying that one should make sure that for each batch an equal number of pairs per class are created and passed to the constrained loss. We can note that given X samples (S$_1$,S$_2$..S$_X$) of a particular class in a batch X$^2$ pairs can be formed of which X are trivial, i.e (S$_i$,S$_i$) as $log$ for these pairs would be 0. We found that forming 2X pairs per class was ideal. However, directly creating 2X pairs per class would lead to an imbalance in class pairs as X varies for each class across each batch. We tackled this problem by only creating a limited number of pairs per class per batch equal to twice the number of samples from the minimum labeled class in that batch. i.e, in a 3 class scenario with counts (X,Y,Z), we form 2*$min$(X,Y,Z) pairs for all classes. We train the entire pipeline for 2000 epochs and found that including the constrained loss from epoch 1000 to be ideal. Including the constrained loss from the beginning led to poor convergence. We used the Adam optimizer with a learning rate of 0.001 for all experiments. The Training algorithm for our pipeline can be found in Algorithm \ref{alg:cap}

\begin{figure*}[t]
    \centering
    \includegraphics[width = 0.7 \linewidth]{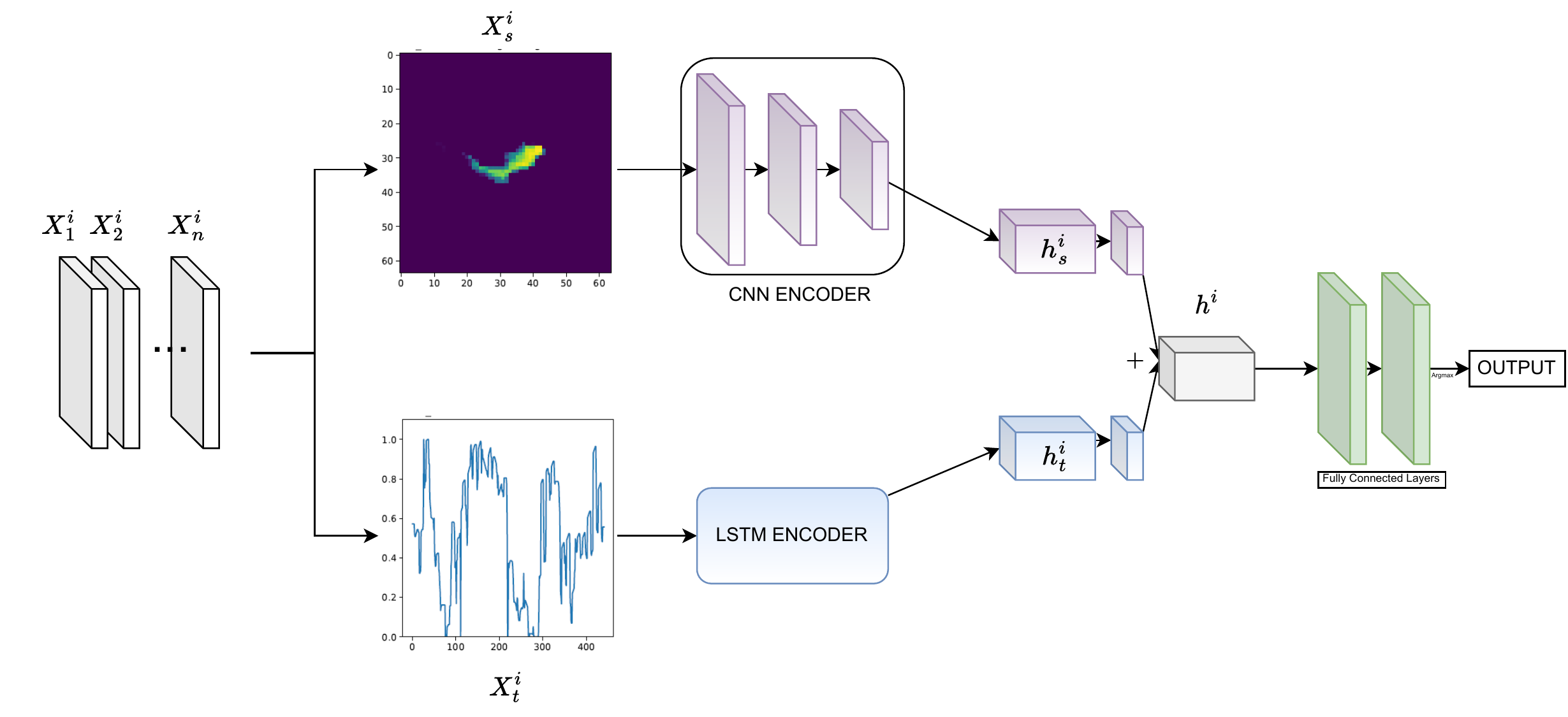}
    \caption{A diagrammatic representation of the architecture of the supervised approach with SLTLAE\_CL used for classification}
    \label{fig:suparchitecture}
\end{figure*}

\begin{algorithm*}[t]
    \caption{Label Augmentation Process}
    \label{alg:aug}
    \begin{algorithmic}
        \Require $N$ Clusters C, Labelled Samples LS, Unlabelled Samples US ($N$ is the number of Clusters)
        \\
        \For {Cluster number i = $1$ to $N$}
            \State Count labelled samples per class in Cluster $C_i \rightarrow (LS_{class1} ,LS_{class2} ,LS_{class3} ,LS_{class4})$
            \State Count number of non zero values Z in $(LS_{class1},LS_{class2},LS_{class3},LS_{class4})$
            
            \If{Z $==$ 1} (Checking if cluster has only one labelled class)
                \State Let the count of that class $i$ that exists in this cluster be $LS_i$
                \If{$LS_i \geq 2$}
                    \State Compute average distance $AVGD$ of $LS_i$'s from the center of cluster $C_i$
                    \For {Unlabelled sample $US$ in Cluster $C_i$}
                        \If{Distance of $US$ from Cluster Center of $C_i$ $\leq AVGD$} 
                            \State Add $US$ as a labelled sample to class $i$
                        \EndIf
                    \EndFor
                \EndIf
            \EndIf        
        \EndFor            
    \end{algorithmic}
\end{algorithm*}

\begin{table*}[t]
    \small
    \caption{F1 classification score to compare representation quality of various methods via training only last two layers with 10 labels per class}
    \label{Table:F1score_lasttwo_update}
    \centering
        \begin{tabular}{|l|cccc|c|}
            \hline
            MODEL                       & Farm      & River     & Stable Lakes  & Seasonal Lakes    & Average Score \\
            \hline
            SAE\_enc\_fixed             & 0.6790    & 0.7626    & 0.4309        & 0.3871            & 0.5649 \\
            TAE\_enc\_fixed             & 0.7854    & 0.5000    & 0.7254        & 0.6926            & 0.6758    \\
            SLTLAE\_enc\_fixed          & 0.8961    & 0.7615    & 0.7005        & 0.6877            & 0.7615 \\
            SLTLAE\_DEC\_enc\_fixed     & 0.8982    & 0.8235    & 0.6839        & 0.6980            & 0.7759\\
            SLTLAE\_CONS\_enc\_fixed    & 0.9203    & 0.8819    & 0.6465        & 0.6802            & 0.7802   \\
            SLTLAE\_CL\_enc\_fixed      & 0.9248    & 0.9085    & 0.7791        & 0.8120            & 0.8512    \\
            \hline
        \end{tabular}
\vspace{-0.5cm}
\end{table*}

\subsection{Extracting new Labels}
\label{Sec:Label Augmentation}
This section shows how we extract new labels from large unlabeled sets and using the encoders from the multimodal autoencoder as pretrained initializations for our supervised classifier. After training SLTLAE\_CL, we cluster the water bodies of the entire training set by applying KMeans on their latent representations, including both the labeled and unlabeled samples. There are several ways to extract new labels, such as taking every sample from clusters with at least one labeled sample or forming tight clusters with only one labeled sample per cluster and then taking all samples within labeled clusters. However, these methods either select too many (which leads to wrong samples being picked up) or too few samples(which leads to loss of valuable samples) from the heterogeneous unlabeled set. To overcome this, we propose a new method of label augmentation, whose detailed procedure can be found in  Algorithm \ref{alg:aug}. Our approach chooses only those samples that lie within the average distance of labeled samples encodings from the center of pure clusters. Our augmentation method eliminates outliers that tend to lie at the edge of clusters and also picks up all valuable samples from the unlabelled set. These extracted new labels are used to augment the limited labels available to train the supervised classifier.

\subsection{Building supervised classifiers}
In limited label scenarios, there are multiple ways to build supervised models: i) Use the limited labels to train a supervised model from scratch ({\bf{\emph{scratch}}}) ii) Train an encoder with the large set of unlabeled and labeled data in an unsupervised/semi-supervised fashion, then, fixing the encoder, add fully connected layers for further training with labeled data ({\bf{\emph{enc\_fixed}}}) iii) Similar to ii), however, instead of keeping the encoder fixed, update the encoder along with the fully connected layers ({\bf{\emph{enc\_upd}}}). In addition to augmenting the existing limited labels, we further transfer the weights of the encoders from our multimodal autoencoder and build the supervised classifier using the few labels to finetune the model in a typical multiclass classification training scenario in one of the three mentioned methods. The diagrammatic representation for the model that would be built using the encoder of SLTLAE\_CL (i.e SLTLAE\_CL\_enc\_fixed or SLTLAE\_CL\_enc\_upd), is shown in Figure \ref{fig:suparchitecture}.


\section{Experimental results}
In this section, we show how the representations produced by our pipeline are superior to other methods and how to leverage these representations to create accurate supervised models. We start by introducing the competitive methods, i.e baselines.
\vspace{-0.3cm}

\subsection{Baselines}
We will build representations using the following methods and then test them in the fixed ({\bf{\emph{enc\_fixed}}}) and update ({\bf{\emph{enc\_upd}}}) classifier setting. 
As competition we to our pipeline we have some baselines to which we compare:
\begin{enumerate}
    \item \textbf{Spatial Autoencoder (SAE)}: This is a spatial-only based autoencoder with the same architecture as that of the spatial part of SLTLAE\_CL. This method takes input as a fraction map and reconstructs the fraction map using the reconstruction loss $\mathcal{L}_{FM\_Rec}$. This is an unsupervised method.
    \item \textbf{Temporal Autoencoder (TAE)}: This is a temporal-only based autoencoder with the same architecture as that of the temporal part of SLTLAE\_CL. This method takes input as time series and reconstructs the time series using the reconstruction loss $\mathcal{L}_{TS\_Rec}$. This is an unsupervised method.
    \item \textbf{Spatial plus Temporal Autoencoder (SLTLAE)}: This is a variation of our pipeline that does not make use of any labels, i.e does not have constrained loss. This method makes use of both modalities and reconstructs both in the exact same fashion as SLTLAE\_CL using the loss $L = \lambda \mathcal{L}_{FM\_Rec} +  \gamma \mathcal{L}_{TS\_Rec}$. This is an unsupervised method.
    \item \textbf{Deep Embedded Clustering Spatial plus Temporal Autoencoder (SLTLAE\_DEC)}: This baseline follows the technique proposed in \cite{xie2016unsupervised}. We maintain the architecture the same as SLTLAE but change the loss components and training procedure as described in \cite{xie2016unsupervised}. This is an unsupervised method.
    \item \textbf{Spatial plus Temporal Autoencoder (SLTLAE\_CONS)}: This baseline follows the algorithm proposed in \cite{zhang2021framework}. Here, we maintain the architecture the same as our pipeline but change the loss components and training procedure as described in \cite{zhang2021framework}. This is a semi-supervised method.
\end{enumerate}

\begin{table*}[t]
    \small
    \caption{F1 Classification score comparison of supervised methods trained in multiple fashions with 10 labels per class}
    \centering
    \label{Table:F1score_all_update}
    \begin{tabular}{|l|cccc|c|}
        \hline
        MODEL & Farm   & River  & Stable Lakes & Seasonal Lakes & Average Score
        \\\hline
        SLTLAE\_scratch & 0.8478 & 0.8339 & 0.5000      & 0.7296 &   0.7278     \\
        \hline 
        SLTLAE\_enc\_fixed      & 0.8961 & 0.7615 & 0.7005      & 0.6877  &   0.7615   \\
        SLTLAE\_DEC\_enc\_fixed      & 0.8982 & 0.8235 & 0.6839      & 0.6980  &    0.7759  \\
        SLTLAE\_CONS\_enc\_fixed     & 0.9203 & 0.8819 & 0.6465 & 0.6802   &  0.7822\\
        SLTLAE\_CL\_ enc\_fixed   & 0.9248 & 0.9085 & 0.7791      & 0.8120  &  0.8561  \\
        \hline 
        SLTLAE\_enc\_upd     & 0.8696 & 0.7302 & 0.8263      & 0.8259   &    0.8130 \\
        SLTLAE\_DEC\_enc\_upd      & 0.9273 & 0.9116 & 0.7071      & 0.7167 &     0.8156  \\
        SLTLAE\_CONS\_enc\_upd     & 0.9266 & 0.9041 & 0.7841     & 0.8134   & 		0.8571	   \\
        SLTLAE\_CL\_enc\_upd  & 0.9479 & 0.9078 & 0.8166      & 0.8333  & 0.8764\\
        \hline
        SLTLAE\_enc\_upd - new labels     & 0.9647 & 0.7775 & 0.7536 & 0.5688 & 0.7662 \\
        SLTLAE\_DEC\_enc\_upd - new labels     & 0.9683 & 0.8764 & 0.7030 & 0.6325 & 0.7951 \\
        SLTLAE\_CONS\_enc\_upd - new labels     &0.9621	& 0.9375 &	0.7746 &	0.7860 & 0.8650\\
        SLTLAE\_CL\_enc\_upd - new labels & 0.9687 & 0.9573 & 0.9710 & 0.9210 & 0.9545 \\
        \hline
    \end{tabular}
\vspace{-0.5cm}
\end{table*}

\begin{table}[t]
    \small
    \caption{Number of New labels extracted by each method using Algorithm \ref{alg:aug}}
    \label{Table:labelaugcount}
    \centering
    \resizebox{\columnwidth}{!}{%
        \begin{tabular}{|l|cccc|}
            \hline
            MODEL           & Farm  & River & Stable Lakes  & Seasonal Lakes\\
            \hline 
            SLTLAE          & 97    & 137   & 47            & 99     \\
            SLTLAE\_DEC     & 94    & 114   & 18            & 45     \\
            SLTLAE\_CONS    & 221   & 57    & 28            & 21\\
            SLTLAE\_CL      & 272   & 110   & 34            & 34  \\
            \hline
        \end{tabular}
    }
\vspace{-0.5cm}
\end{table}

\subsection{Comparison of Representation quality}
To compare the quality of representations produced by our proposed scheme and baselines, we employ a strategy similar to that used by Asano et al. ~\cite{selflabellingiclr}. We compare models built in the enc\_fixed approach by varying the encoders produced by our scheme and the baselines. The intuition behind training such a model in this fashion is that if the encodings given by the encoder are good, the fully connected layers could easily learn a classification boundary using the limited labels given to it, which would lead to better test classification scores.

For this setup, we chose to have only ten labels from each class present, leading to 40 labels, with the rest of the training set being unlabelled. First, we get the encoder weights from SAE, TAE, SLTLAE, SLTLAE\_DEC, SLTLAE\_CONS, and SLTLAE\_CL that were obtained using the unsupervised/semi-supervised training described before. SLTLAE\_CONS and SLTLAE\_CL use the 40 labels by creating pairs and applying constraints. Finally, we add the two fully connected layers to create the respective enc\_fixed model and train the network using the 40 train labels for 50 epochs using Cross entropy loss. During this training step, the encoder stays fixed, and only the weights of two fully connected layers change. After this training is over, using each model, we classify each test sample into one of the four classes reporting the F1 score table for test classification results in Table \ref{Table:F1score_lasttwo_update}

From the table, we can observe that SLTLAE\_enc\_fixed  outperforms both SAE\_enc\_fixed  and TAE\_enc\_fixed in almost every class, which shows the power of combining modalities to get better representations. Further, SLTLAE\_DEC\_enc\_fixed performs slightly better than SLTLAE\_enc\_fixed, which can be attributed to the embedded clustering aspect of SLTLAE\_DEC. We can also observe that models built using encoders from semisupervised methods perform better than the unsupervised competitors, showing that constrained loss is helping in improving representations. However, it is essential to note the difference in performance between SLTLAE\_CONS\_enc\_fixed and SLTLAE\_CL\_enc\_fixed, SLTLAE\_CL\_enc\_fixed performs better than SLTLAE\_CONS\_enc\_fixed in all classes with significant improvements in Stable Lakes and Seasonal Lakes. The results show that our proposed constrained loss leads to better representations in limited-label scenarios.

\subsection{Training a Supervised Model with few labels}
This section compares supervised methods built in the \emph{enc\_upd} fashion to those built using \emph{enc\_fixed}. This section will explore which fashion yields the most accurate supervised model. We also maintained the same settings as the previous section: 10 labels from each class are present, with the rest of the training set being unlabelled. We also saw from Table \ref{Table:F1score_lasttwo_update} that SAE and TAE do not lead to good representations, so we exclude them from the analysis here and focus only on dual modality-based methods. The top part of Table \ref{Table:F1score_all_update} reports the F1 scores on the test set for each baseline and our pipeline trained in the three ways described before. One point to note is that SLTLAE, SLTLAE\_DEC, SLTLAE\_CONS, and SLTLAE\_CL have the same \emph{scratch} model and so all share the SLTLAE\_scratch entry in the Table. The \emph{enc\_fixed} model scores are imported from Table \ref{Table:F1score_lasttwo_update} as the experiment setting is the same.

We can observe from the scores in Table \ref{Table:F1score_all_update} that \emph{scratch} model is not able to perform too well and also that \emph{enc\_upd} model performs better than its \emph{enc\_fixed} counterpart. We can see that SLTLAE\_CL\_enc\_upd performs the best out of the baselines, including SLTLAE\_CONS\_enc\_upd, showing two things: (i) \emph{enc\_upd} is the best way to get a supervised model in limited-label scenario and (ii) the encoder provided by SLTLAE\_CL was the best starting point to build this model. We believe that this is due to the ability of our proposed approach to creating a better representation that also makes it possible to be more effective in augmenting labels. We also did another experiment in which we used 70 labels from each class. As expected, all results improve and become closer (although relative ordering remains the same). This shows that our approach is particularly suitable for limited-label scenarios. (Results not shown due to space constraints).

\subsection{Augmenting labels from unlabeled set}
As a final experiment to show the power of SLTLAE\_CL, we delve into the case of label augmentation from the large pool of unlabeled data present in our dataset. Similar to the previous experiments, we use ten labels per class to train SLTLAE, SLTLAE\_DEC,  SLTLAE\_CONS, and SLTLAE\_CL in their respective unsupervised/semisupervised approach. After training, we cluster the entire training set, including the labeled samples, using KMeans clustering into 50 clusters, with the ten labeled samples lying in some clusters. We then use Algorithm\ref{alg:aug} to add labels to each of the four classes in all 4 cases.

Using these new labels and the ten labels from each class, we trained a supervised approach via \emph{enc\_upd} method and tested the performance on the test set with the classification results shown in the bottom half of Table \ref{Table:F1score_all_update}. We report the number of new samples picked up by each baseline and our approach for each class in Table \ref{Table:labelaugcount}. From the Table, we can observe that the performance of SLTLAE\_enc\_upd - new labels, SLTLAE\_DEC\_enc\_upd - new labels, and SLTLAE\_CONS\_enc\_upd - new labels is significantly worse than SLTLAE\_CL\_enc\_upd - new labels, and it is even more interesting to see that the performance of some classes using SLTLAE\_enc\_upd is, in fact, better than that of the model trained using the new labels. We see a similar drop in performance for some classes in SLTLAE\_CONS as well, even though it is a semisupervised method. These two observations show that not only are the new samples picked up by SLTLAE\_CL better but also that the new samples picked up by SLTLAE, SLTLAE\_DEC and SLTLAE\_CONS are not accurate and are hurting performance. This again shows how our paradigm is much better than standard methods and how one can extract new labels from large unlabelled datasets using our pipeline. 

\vspace{-0.3cm}
\section{Conclusion}
This paper proposes a semi-supervised spatiotemporal approach to building better representations for large unstructured datasets. We showed how combining different data modalities can help build improved representations. We also showed how with a few labels, one could improve the quality of representations using a constrained loss framework. We then finally explored how our method was effective for label augmentation from large pools of unlabelled data, which was verified by building supervised around the new labels, leading to an increase in accuracy of about 10\% across four classes of farms, rivers, stable lakes, and seasonal lakes. We believe this feature of our pipeline can potentially pave the way for the extraction of labels when new classes are added to large unexplored datasets. Another point to note is that our paradigm can also be used with any manually defined classes. For example, if someone chose to add a new class to ReaLSAT, they could run the pipeline and extract new labels for that new class. We believe this is a handy feature, especially with datasets like ReaLSAT that are in the nascent state of exploration, where new classes appear as more analysis is done. 
\vspace{-0.4cm}
\bibliographystyle{plain}
\bibliography{bibliography}

\end{document}